\documentclass[11pt]{article}
\usepackage{colacl}
\title{Comparing two trainable grammatical relations finders
\thanks{This paper reports on work performed at the MITRE Corporation
under the support of the MITRE Sponsored Research Program. Marc Vilain, Lynette Hirschman and Warren Greiff have helped make this work happen. Christine Doran and John Henderson provided helpful editing. Copyright \copyright 2000 The
MITRE Corporation. All rights reserved.}}

\author{Alexander Yeh\\Mitre Corp.\\202 Burlington Rd.\\Bedford, MA 01730\\USA\\asy@mitre.org}

\begin{document}

\maketitle


\begin{abstract}
{\begin{picture}(0,0)
\put(0,200){Appears in the 18th International Conference on Computational Linguistics (COLING 2000),}
\put(10,185){pages 1146-1150, Saarbr\"{u}cken, Germany, July, 2000.}
\put(0,-505){cs.CL/0008004}
\end{picture}}
Grammatical relationships (GRs) form an important level of natural
language processing, but different sets of GRs are useful for
different purposes. Therefore, one may often only have time to obtain
a small training corpus with the desired GR annotations. On such a small
training corpus, we compare two systems. They use different learning
techniques, but we find that this difference by itself only has a
minor effect.
A larger factor is that in English, a
different GR length measure appears better suited for finding simple
argument GRs than for finding modifier GRs.  We also find that partitioning
the data may help memory-based learning.
\end{abstract}

\section{Introduction}

Grammatical relationships (GRs), which include arguments (e.g.,
subject and object) and modifiers, form an important level of natural
language processing. GRs in the sentence 
\begin{center}
{\em Yesterday, my cat ate the food in the bowl.} 
\end{center}
include {\em ate} having the subject \mbox{\em my cat}, the object
\mbox{\em the food} and the time modifier {\em Yesterday}, and
\mbox{\em the food} having the location modifier \mbox{\em in (the
bowl)}.

However, different sets of GRs are useful for different purposes. For
example, \newcite{FVY99} is interested in semantic interpretation, and
needs to differentiate between time, location and other modifiers. The
SPARKLE project \cite{Sparkle1}, on the other hand, does not
differentiate between these types of modifiers. As has been mentioned
by John Carroll (personal communication), this is fine for information
retrieval. Also, having less differentiation of the modifiers can make
it easier to find them \cite{FVY99}.

Unless the desired set of GRs matches the set already annotated in
some large training corpus (e.g., the \newcite{BVD99} GR finder used
the GRs annotated in the Penn Treebank \cite{PennTreebank}), one will
have to either manually write rules to find the GRs or annotate a
training corpus for the desired set. Manually writing rules is
expensive, as is annotating a large corpus. We have performed
experiments on learning to find GRs with just a small annotated
training set. Our starting point is the work described in
\newcite{FVY99}, which used a fairly small training set.

This paper reports on a comparison between the transformation-based
error-driven learner described in \newcite{FVY99} and the memory-based
learner for GRs described in \newcite{BVD99} on finding GRs to
verbs\footnote{That is, GRs that have a verb as the relation
target. For example, in {\em Cats eat.}, there is a ``subject''
relation that has {\em eat} as the target and {\em Cats} as the
source.} by retraining the memory-based learner with the data used
in \newcite{FVY99}. We find that the transformation versus
memory-based difference only seems to cause a small difference in the
results.  Most of the result differences seem to instead be caused by
differences in the representations and information used by the
learners. An example is that different GR length measures are
used. In English, one measure seems better for recovering simple
argument GRs, while another measure seems better for modifier GRs.  We
also find that partitioning the data sometimes helps memory-based
learning.

\section{Differences Between the Two Systems}

\newcite{FVY99} and \newcite{BVD99} both describe learning systems to
find GRs. The former (TR) uses transformation-based error-driven
learning \cite{BandR94} and the latter (MB) uses memory-based learning
\cite{TiMBL}.

In addition, there are other differences. The TR system includes
several types of information not used in the MB system (some because
memory-based systems have a harder time handling set-valued
attributes): possible syntactic (Comlex) and semantic (Wordnet)
classes of a chunk headword, the stem(s) and named-entity category
(e.g., person, location), if any, of a chunk headword, lexemes in
a chunk besides the headword, pp-attachment estimate and certain verb
chunk properties (e.g., passive, infinitive).

Some lexemes (e.g., coordinating conjunctions and punctuation) are
usually outside of any chunk. The TR system will store these in an
attribute of the nearest chunk to the left and to the right of such a
lexeme. The MB system represents such lexemes as if they are one word
chunks. The MB system cannot use the TR system method of storage because
memory-based systems have difficulties with set-valued attributes
(value is 0 or more lexemes).

The MB system (and not the TR system) also examines the number of commas
and verb chunks crossed by a potential GR.

The space of possible GRs searched by the two systems is slightly
different. The TR system searches for GRs of length three chunks
or less. The MB system searches for GRs which cross at most
either zero (target to the source's left) or one (to the right) verb
chunks.

Also, slightly different are the chunks examined relative to a
potential GR. Both systems will examine the target and source chunks,
plus the source's immediate neighboring chunks. The MB system also
examines the source's second neighbor to the left. The TR system
instead also examines the target's immediate neighbors and all the
chunks between the source and target.

The TR system has more data partitioning than the MB system. With the TR
system, possible GRs that have a different source chunk type (e.g.,
noun versus verb), or a different relationship type (e.g., subject
versus object) or direction or length (in chunks) are always
considered separately and will be affected by different rules. The MB
system will note such differences, but may decide to ignore some or
all of them.

\section{Comparing the Two Systems}
\subsection{Experiment Set-Up}

One cannot directly compare the two systems from the descriptions
given in \newcite{FVY99} and \newcite{BVD99}, as the results in the
descriptions were based on different data sets and on different
assumptions of what is known and what needs to be found.

Here we test how well the systems perform using the same small
annotated training set, the 3299 words of elementary school reading
comprehension test bodies used in \newcite{FVY99}.\footnote{Note that
if we had been trying to compare the two systems on a large annotated
training set, the MB system would do better by default just because the
TR system would take too long to process a large training set.} We are
mainly interested in comparing the parts of the system that takes in
syntax (noun, verb, etc.) chunks (also known as groups) and finds the
GRs between those chunks. So for the experiment, we used the general
TiMBL system \cite{TiMBL} to just reconstruct the part of the MB system
that takes in chunks and finds GRs. The input to both this
reconstructed part and the TR system is data that has been manually
annotated for syntax chunks and GRs, along with automatic lexeme and
sentence segmentation and part-of-speech tagging. In addition, the TR
system has manual named-entity annotation, and automatic estimations
for verb properties and preposition and subordinate conjunction
attachments \cite{FVY99}. Because the MB system was originally designed
to handle GRs attached to verbs (and not noun to noun GRs, etc.), we
ran the reconstructed part to only find GRs to verbs, and ignored
other types of GRs when comparing the reconstructed part with the TR
system. The test set is the 1151 word test set used in
\newcite{FVY99}. Only GRs to verbs were examined, so the effective
training set GR count fell from 1963 to 1298 and test set GR count
from 748 to 500.

\subsection{Initial Results}\label{ss:rule-vs-case-results}

In looking at the test set results, it is useful to divide up
the GRs into the following sub-types:
\begin{enumerate}
\item \label{en:sim-arg} Simple arguments: subject, object, indirect
object, copula subject and object, expletive subject (e.g., {\em ``It''}
in {\em ``It rained today.''}).

\item \label{en:mod} Modifiers: time, location and other modifiers.

\item \label{en:messy-arg} Not so simple arguments: arguments that
syntactically resemble modifiers. These are location objects, and also
subjects, objects and indirect objects that are attached via a
preposition.
\end{enumerate}

Neither system produces a spurious response for
type~\ref{en:messy-arg} GRs, but neither system recalls many of the
test keys either. The reconstructed MB system recalls 6 of the 27 test
key instances (22\%), the TR system recalls 7 (26\%).  A possible
explanation for these low performances is the lack of training
data. Only 58 (3\%) of the training data GR instances are of this
type.

The type~\ref{en:mod} GRs are another story. There are 103 instances of
such GRs in the test set key. The results are
\begin{center}
\begin{tabular}{|l|r|r|r|}\hline
\multicolumn{4}{|c|}{Type~\ref{en:mod} GRs}\\
System & Recall & Precision & F-score \\ \hline
MB &  47 (46\%) & 49\% & 47\% \\ \hline
TR &  25 (24\%) & 64\% & 35\% \\ \hline
\end{tabular}
\end{center}
Recall is the number (and percentage) of the keys that are recalled.
Precision is the number of correctly recalled keys divided by the
number of GRs the system claims to exist. F-score is the harmonic mean
of recall ($r$) and precision ($p$) percentages. It equals
$2pr/(p+r)$. Here, the differences in $r$, $p$ and F-score are all
statistically significant.\footnote{When comparing differences in this
paper, the statistical significance of the higher score being better
than the lower score is tested with a one-sided test. Differences
deemed statistically significant are significant at the 5\% level.
Differences deemed non-statistically significant are not significant
at the 10\% level.}  The MB system performs better as measured
by the F-score. But a trade-off is involved. The MB system has both a
higher recall and a lower precision.

The bulk (370 or 74\%) of the 500 GR key instances in the test set are
of type~\ref{en:sim-arg} and most of these are either subjects or
objects. With type~\ref{en:sim-arg} GRs, the results are
\begin{center}
\begin{tabular}{|l|r|r|r|}\hline
\multicolumn{4}{|c|}{Type~\ref{en:sim-arg} GRs}\\
System & Recall & Precision & F-score \\ \hline
MB &  231 (62\%) & 66\% & 64\%  \\ \hline
TR &  284 (77\%) & 82\% & 79\% \\ \hline
\end{tabular}
\end{center}
With these GRs, the TR system performs considerably better
both in terms of recall and precision. The differences in all three
scores are statistically significant.

Because 74\% of the GR test key instances are of
type~\ref{en:sim-arg}, where the TR system performs
better, this system performs better when looking at the results for
all the test GRs combined. Again, all three score differences are
statistically significant:
\begin{center}
\begin{tabular}{|l|r|r|r|}\hline
\multicolumn{4}{|c|}{Combined Results}\\
System & Recall & Precision & F-score \\ \hline
MB &  284 (57\%) & 63\% & 60\%  \\ \hline
TR &  316 (63\%) & 80\% & 71\% \\ \hline
\end{tabular}
\end{center}

Later, we tried some extensions of the reconstructed MB system to try
to improve its overall result. We could improve the overall result by
a combination of using the {\em IB1} search algorithm (instead of {\em
IGTREE}) in TiMBL, restricting the potential GRs to those that crossed
no verb chunks, adding estimates on preposition and complement
attachments (as was done in TR) and adding information on verb chunks
about being passive, an infinitive or an unconjugated present
participle. The overall F-score rose to 65\% (63\% recall, 67\%
precision). This is an improvement, but the TR system is still
better. The differences between these scores and the other MB and TR
combined scores are statistically significant.

\subsection{Exploring the Result Differences}

\subsubsection{Type~\ref{en:mod} GRs: modifiers}
The reconstructed MB system performs better at type~\ref{en:mod}
GRs. How can we account for this result difference?

Letting the TR system find longer GRs (beyond 3 chunks in length) does
not help much. It only finds one more type~\ref{en:mod} GR in
the test set (adds 1\% to recall and 1\% or less to precision).

Rerunning the TR system rule learning with an information organization
closer to the MB system produces the same 47\% F-score as the MB system
(recall is lower, but precision is higher). Specifically, we got this
result when the TR system was rerun with no information on
pp-attachments, verb chunk properties (e.g., passive, infinitive),
named-entity labels or headword stems. Also, the TR system now examines
the chunks examined by the original MB system: target, source and
source's neighbors.  In addition, instead of 6 absolute length
categories (target is 3 chunks to the left, 2 chunks, 1 chunk, and
similarly for the right), the GRs considered now just fall into and
are partitioned into 3 relative categories: target is the first verb
chunk to the left, similarly to the right and target is the second
verb chunk to the right. The MB system can distinguish between these
same relative categories.

Redoing this TR system rerun {\em without} chunk headword syntactic or
semantic classes produces a 46\% F-score. If in addition, the
pp-attachment, verb chunk property, named-entity label and headword
stem information are added back in, the F-score actually drops to
43\%.  The differences between these 47\%, 46\% and 43\% rerun scores
are not statistically significant.

So with type~\ref{en:mod} GRs, MB system's better performance seems to
be mainly due its ability to differentiate the potential GRs by the
feature of the number of verb chunks crossed by a GR. In particular,
making this and a few other changes to the TR system increases its
F-score to the MB system's F-score, and the other changes (removing
certain information) does not have a significant effect. So using the
right features can make a large difference.

For these type~\ref{en:mod} GRs (modifiers) in English, it does seem
that the number of verb chunks crossed is a better way to group
possible modifiers than the absolute chunk length. An example is
comparing \mbox{\em I fly on Tuesday.}  and \mbox{\em I fly home from
here on Tuesday.} In both sentences, \mbox{\em on Tuesday} is a time
modifier of {\em fly} and {\em on} crosses no verbs to reach {\em fly}
({\em on} attaches to the first verb to its left). But in the first
sentence {\em on} is next to {\em fly}, while in the second sentence,
there are three chunks separating {\em on} and {\em fly}.

\subsubsection{Type~\ref{en:sim-arg} GRs: simple arguments}

For type~\ref{en:sim-arg} GRs, the TR system performs better. How can
we account for this?

Much of the extra information the TR system examines (compared to the MB
system) does not seem to have much of an effect. When the TR system was
rerun with no information on headword syntactic or semantic classes,
named-entity labels or headword stems, the F-score increased from 79\%
to 80\%. Another rerun that in addition had no information on
pp-attachment estimates or any of the non-headwords in the chunks also
had an F-score of 80\%. A third rerun that furthermore had no
information on verb chunk properties (e.g., passive, infinitive) had
an F-score of 78\%. In this set of F-scores, only the differences
between the 80\% scores and the 78\% score are statistically
significant.

Some MB system reruns showed factors that seemed to matter more.
In the first rerun, we partitioned the data by potential GR source
chunk type (e.g, noun versus verb) and ran a separate memory-based
training and test for each partition. The combined F-score increased
from 64\% to 69\%. Afterwards, we made a rerun that resembled the TR
system run with the 78\% F-score (except that memory-based learning
was used): only GRs of length 3 chunks or less were considered, the
data was partitioned (in addition to source chunk type) by GR length
and direction (e.g., target is two chunks to the left) and also by
relation type (separate runs for each type), the comma and verb chunk
crossing counts were not considered, and the chunks normally examined
by the TR system were examined. This further increased the F-score to
75\%. In this set of F-scores, all the differences are statistically
significant and all the F-scores in this set are statistically
significantly different from the TR system runs with the 78\% and 79\%
F-scores.

From the statistically significant score differences, it seems that
partitioning data by potential GR source chunk type helps (increase
from 64\% to 69\%), as does the rest of the partitioning performed and
making some slight changes in what is examined (increase to 75\%),
using transformation-based learning instead of memory-based learning
(increase to 78\%) and using verb chunk property information (increase
to 80\%).

In the original MB system run, the source chunk type and the potential
GR length and direction were already determined by the memory-based
learner to be the most important attributes examined. So why would
partitioning the data and runs by the values of these attributes be of
extra help?  A possible answer is that for different values, the
relative order of importance of the other attributes (as determined by
the memory-based learner) changes. For example, when the source chunk
type is a noun, the second most important attribute is the source
chunk's headword when the target is one to the right, but is the
source chunk's right neighbor's headword when the target is one to the
left. Partitioning the data and runs lets these different relative
orders be used. Having one combined data set and run means that only
one relative order is used. Note that while this partitioning may not
be the standard way of using memory-based learning, it is consistent
with the central idea in memory-based learning of storing all the
training instances and trying to find the ``nearest'' training
instance to a test case.

Another question is why using transformation-based (rule) learning
seems to be slightly better than memory-based learning for these
type~\ref{en:sim-arg} GRs. Memory-based learning keeps all of the
training instances and does not try to find generalizations such as
rules \cite[Ch.~4]{TiMBL}. However, with type~\ref{en:sim-arg} GRs, a
few simple generalizations can account for many of the instances. In
the manner of \newcite{Stevenson98}, we wrote a set of six simple
rules that when run on the test set type~\ref{en:sim-arg} GRs produces
an F-score of 77\%. This is better than what our reconstructed MB
system originally achieved and is close to the TR system's original
results (close enough not to be statistically significantly
different). An example of these six rules: IF (1) the center chunk is
a verb chunk and (2) is not considered as possibly passive and (3) its
headword is not some form of {\em to be} and (4) the right neighbor is
a noun or verb chunk, 
THEN consider that chunk to the right as being an object of the center
chunk.


\section{Discussion}

GRs are important, but different sets of GRs are useful for different
purposes. We have been looking at ways of improving automatic GR
finders when one has only a small amount of data with the desired GR
annotations. In this paper, we compared a transformation rule-based
system with a memory-based system on a small training corpus. We found
that on GRs that point to verbs, most of the result differences can be
accounted for by differences in the representations and information
used. The type of GR determines which information is more important.
The rule versus memory-based difference itself only seems to produce a
small result difference. We also find that partitioning the data may
help memory-based learning.



\end{document}